\documentclass[10pt,twocolumn,letterpaper]{article}

\usepackage{cvpr}              

\usepackage{colortbl}
\usepackage{amssymb}
\usepackage{pifont}
\usepackage{booktabs,multirow}
\usepackage{makecell}
\usepackage{tabulary}
\usepackage{fontawesome5}
\usepackage{bbding}

\newlength\savewidth\newcommand\shline{\noalign{\global\savewidth\arrayrulewidth
  \global\arrayrulewidth 1pt}\hline\noalign{\global\arrayrulewidth\savewidth}}

\usepackage{amsmath}
\usepackage{capt-of}
\usepackage[dvipsnames]{xcolor}
\usepackage{xspace}
\usepackage{enumitem}
\usepackage{amsmath,amssymb,amsbsy,amsfonts,dsfont,pifont,bm,bbm,mathrsfs,mathtools,nicefrac}
\usepackage{algorithm,algpseudocode,listings}
\usepackage{booktabs,multirow,adjustbox,diagbox,threeparttable,tabularray}

\newcommand{\benchmarkname}{CoDanceBench}
\newcommand{\benchmark}{\texttt{\benchmarkname}\xspace}

\newcommand{\methodname}{CoDance}
\newcommand{\method}{\texttt{\methodname}\xspace}

\newcommand{\tocite}[1]{{\color{red} [TO CITE]}}

\definecolor{CQColor}{rgb}{0.0,0.0,1.0} 

\definecolor{TSColor}{rgb}{0.5,0.0,0.8} 

\definecolor{CQRColor}{rgb}{1.0,0.0,1.0} 

\definecolor{cvprblue}{rgb}{0.21,0.49,0.74}
\usepackage[pagebackref,breaklinks,colorlinks,allcolors=cvprblue]{hyperref}
\usepackage{wrapfig}
\usepackage[capitalize]{cleveref}  
\crefname{section}{Sec.}{Secs.}
\Crefname{section}{Section}{Sections}
\crefname{table}{Tab.}{Tabs.}
\Crefname{table}{Table}{Tables}
\crefname{figure}{Fig.}{Figs.}
\Crefname{figure}{Figure}{Figures}
\crefname{equation}{Eq.}{Eqs.}
\Crefname{equation}{Equation}{Equations}
\definecolor{baseColor}{rgb}{0.75,0.05,0.1}

\definecolor{checkmarkColor}{rgb}{0.1,0.75,0.1}

\definecolor{demphcolor}{RGB}{144,144,144}

\def\paperID{1233} 
\def\confName{CVPR}
\def\confYear{2026}

\title{CoDance: An Unbind-Rebind Paradigm for Robust Multi-Subject Animation}

\author{Shuai Tan$^{1,2}$\footnotemark[1]\>\,, Biao Gong$^{2}$\footnotemark[2]\>\,\footnotemark[3]\>\,, Ke Ma$^{3}$, Yutong Feng$^{4}$, Qiyuan Zhang$^{2}$,\\ 
Yan Wang$^{5}$, Yujun Shen$^{2}$, Hengshuang Zhao$^{1}$\footnotemark[3]\>\,\\[5pt]
{$^1$The University of Hong Kong\ \ $^2$Ant Group\ \  $^3$Huazhong University of Science and Technology} \\
{$^4$Tsinghua University\ \ $^5$University of North Carolina at Chapel Hill}\\
}

\begin{document}

\twocolumn[{
\maketitle
\begin{center}
    \vspace{-2pt}
    \includegraphics[width=1\linewidth]{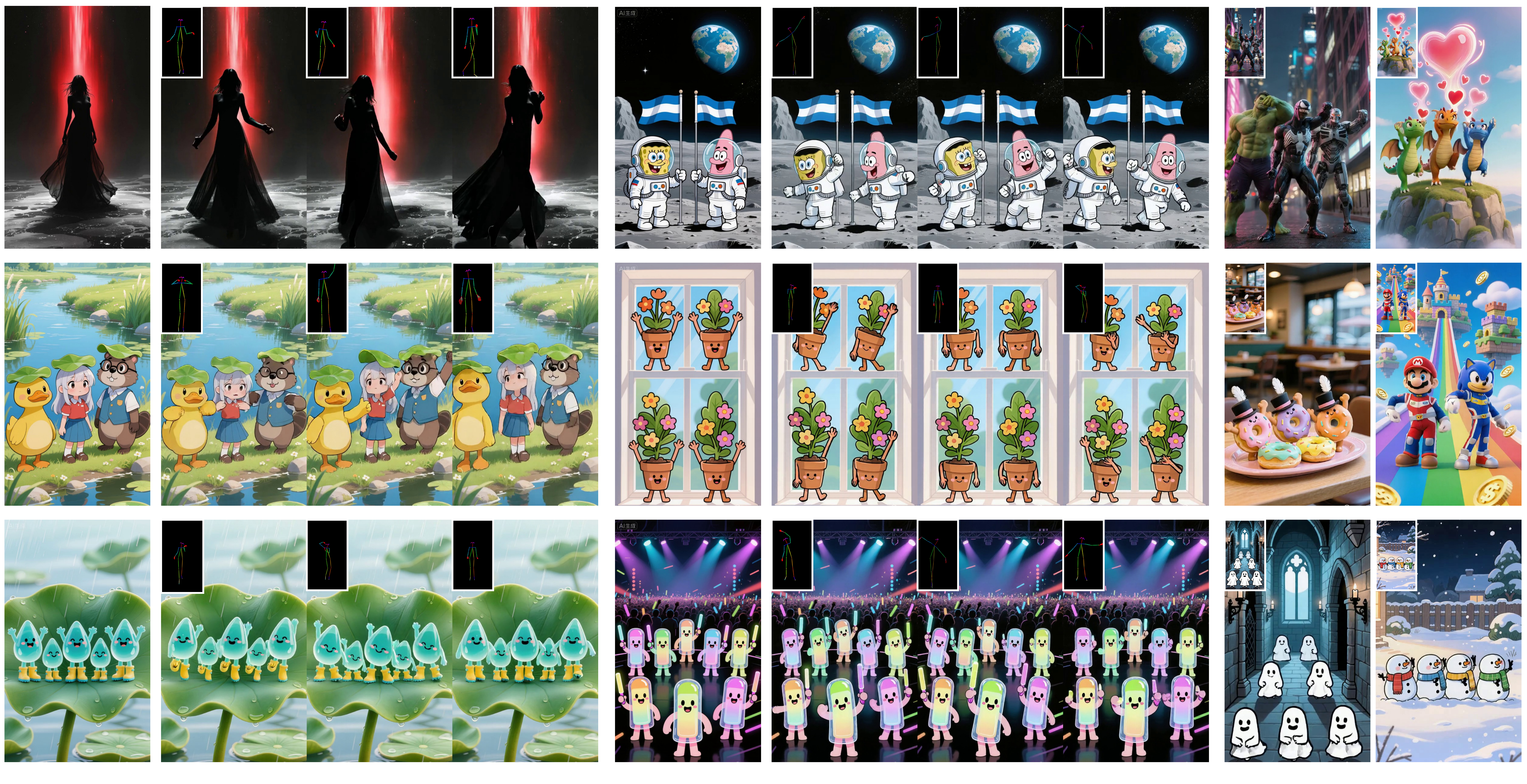}
        \vspace{-10pt}
    \captionof{figure}{ \textbf{Multi-subject animations generated by \method.} Given a single (potentially misaligned) driving pose sequence and one multi-subject reference image, \method produces coordinated, pose-controllable \emph{group dances} without per-subject spatial alignment.}
    \label{fig:teaser}
\end{center}
}]

\footnotetext[1]{Work done during internship at Ant Group. \footnotemark[2]Project lead.}
\footnotetext[3]{Corresponding author.}

\begin{figure*}[h]
   \begin{center}
    \includegraphics[width=1\linewidth]{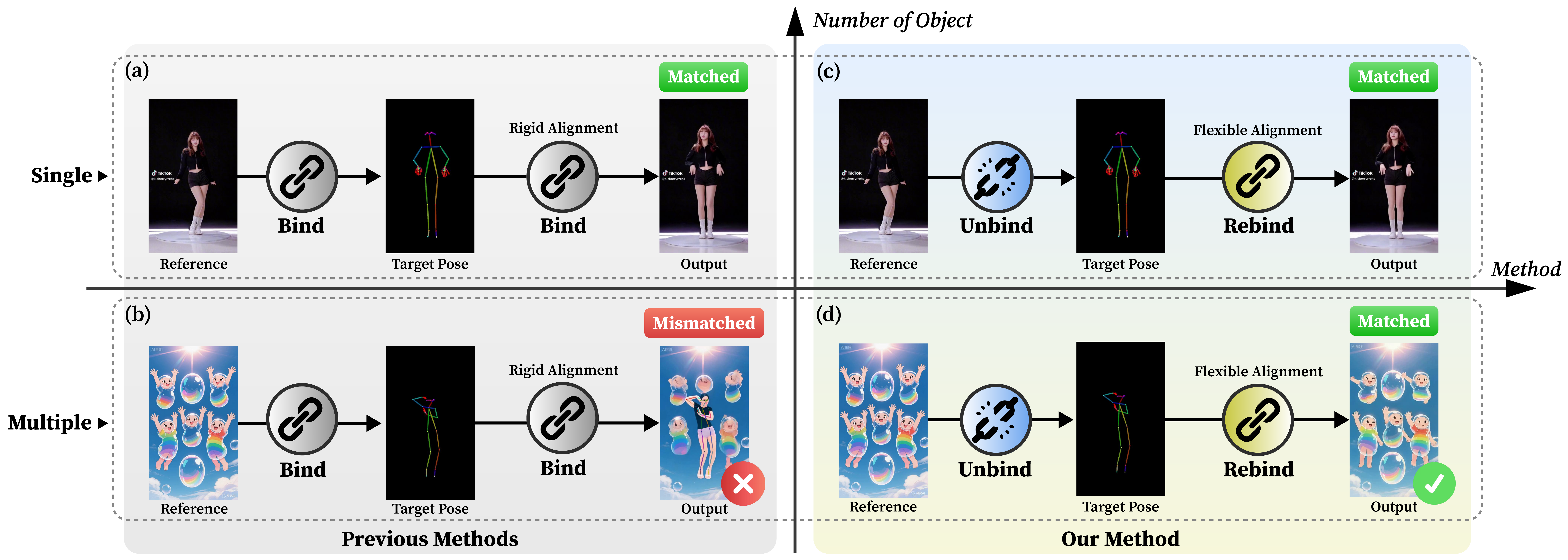}
    \end{center}
    
    \caption{\textbf{The illustration of \method motivation.}
    Although excelling at single-person animation, prior methods fail when handling multiple subjects due to a rigid binding between the reference and target pose, which results in mismatched outputs. By contrast, our Unbind-Rebind method successfully decouples motion from appearance, yielding compelling results.
    }
    \label{fig:introduction}
    \vspace{-3mm}
\end{figure*}

\begin{abstract}
Character image animation is gaining significant importance across various domains, driven by the demand for robust and flexible multi-subject rendering. While existing methods excel in single-person animation, they struggle to handle arbitrary subject counts, diverse character types, and spatial misalignment between the reference image and the driving poses. We attribute these limitations to an overly rigid spatial binding that forces strict pixel-wise alignment between the pose and reference, and an inability to consistently rebind motion to intended subjects. To address these challenges, we propose \method, a novel Unbind-Rebind framework that enables the animation of arbitrary subject counts, types, 
and spatial configurations conditioned on a single, potentially misaligned pose sequence.
Specifically, the Unbind module employs a novel pose shift encoder to break the rigid spatial binding between the pose and the reference by introducing stochastic perturbations to both poses and their latent features, thereby compelling the model to learn a location-agnostic motion representation. To ensure precise control and subject association, we then devise a Rebind module, leveraging semantic guidance from text prompts and spatial guidance from subject masks to direct the learned motion to intended characters. Furthermore, to facilitate comprehensive evaluation, we introduce a new multi-subject \benchmark. Extensive experiments on \benchmark and existing datasets show that \method achieves SOTA performance, exhibiting remarkable generalization across diverse subjects and spatial layouts. The code and weights will be open-sourced.  Project page: \href{https://lucaria-academy.github.io/CoDance/}{https://lucaria-academy.github.io/CoDance/}
\vspace{-5mm}
\end{abstract}

\section{Introduction}
\label{sec:intro}

Character Image Animation aims to generate lifelike, high-fidelity videos from a reference image and a target pose sequence~\cite{Animateanyone}. Compared to the well-studied field of single-person animation, multi-subject ($\geq$2) scenarios~\cite{multidance1} are more prevalent and valuable in practice, such as ensemble performance, and a wide range of applications including advertising and educational content creation~\cite{guo2023animatediff, magicanimate, tu2025stableanimator}.
As shown in Fig.~\ref{fig:introduction}, although recent methods~\cite{unianimatedit, wang2024unianimate, tu2025stableanimator, tu2025stableanimator++, mimicmotion2024, musepose,peng2024controlnext} have achieved impressive progress in single-person animation, most of them still struggle to handle multi-subject scenarios~\cite{multidance2}, leading to the generation of spurious or entangled subjects. Consequently, achieving robust character animation under arbitrary subject counts and spatial layouts remains a pressing problem.

Recently, there have been promising efforts~\cite{multidance2, multidance1} toward two-person animation. For instance, Follow-Your-Pose-V2~\cite{multidance1} operates in an image-to-video paradigm, leveraging multi-branch control modules to generate multi-animation videos. Despite the success in two-person cases, they still face three key limitations in general settings: (1) Limited scalability in subject count. The rigid pose-conditioned binding constrains the control and fusion pipeline, preventing extension beyond two subjects. (2) Positional sensitivity. They impose stringent constraints on initial subject locations, requiring perfect spatial alignment between the reference image and target pose. Mild misalignment often leads to failure induced by over-forcing alignment. (3) Restricted subject types. Current methods are typically optimized for real humans and generalize poorly to subjects that deviate from humans (\textit{e.g.}, anthropomorphic cartoons)~\cite{animatex}. These limitations severely constrain the practical applicability of current methods. 

We attribute these issues primarily to two factors: (1) as illustrated in Fig.~\ref{fig:introduction} (b), an excessively rigid spatial \textit{binding} between the pose skeleton and the reference subjects, which forces strict pixel-wise alignment; and (2) the lack of an explicit focus mechanism (\textit{e.g.} `Rebind' in Fig.~\ref{fig:introduction} (d)) to reliably isolate and animate the intended target. 
Collectively, previous methods fail to infer the desired motion from target poses and to correctly localize the corresponding subjects in the reference image.

To address these issues, we present \method, which animates arbitrary subject types, counts, and spatial positions, given a single pose sequence that is not spatially aligned with the reference image. Our key insight is twofold: (1) improve the flexibility of model in understanding motion semantics rather than equating poses with pixel-wise alignment; and 
(2) to equip the model with autonomous alignment and explicit focus mechanisms, enabling it to accurately localize and animate the intended subjects even under spatial misalignment.
These insights motivate our Unbind and Rebind paradigm, respectively. Concretely, as presented in Fig.~\ref{fig:introduction} (d), we begin with Unbind: during training, we apply scale and shift perturbations to both the explicit pose skeletons and their latent features, deliberately weakening the strong spatial binding between pose and reference image while preserving motion semantics. 
This suppresses the pixel-alignment shortcut and encourages the model to learn subject-agnostic motion semantics and temporal coherence, thereby improving robustness to pose-induced layout shifts and enhancing flexibility in motion understanding.

However, while unbinding relaxes the strict spatial coupling between the pose and the reference image, enabling the model to grasp motion semantics, it leaves the model unable to localize the subjects that should be animated. Therefore, we introduce Rebind, which restores precise control from both semantic and spatial perspectives. Semantically, we add a text branch~\cite{t5} that explicitly specifies the identity and number of subjects to be animated. To mitigate the scarcity and limited diversity of textual labels in animation datasets, we propose a mixed-data training strategy that involves jointly training with large, readily available text-to-video data to strengthen textual understanding and semantic binding. Spatially, we employ an offline segmentation model (\textit{e.g.}, SAM~\cite{sam2}) on the reference image to obtain high-quality subject masks, which serve as external conditions to rebind pose and target subjects in space and explicitly confine the action region. Through this Unbind–Rebind mechanism, the model acquires both elastic motion semantics and accurate target localization under misalignment. We further construct a multi-subject benchmark \benchmark to comprehensively evaluate our approach. Experiments show consistent improvements across multiple quantitative and perceptual metrics, validating the effectiveness of \method.

Our contributions are summarized as follows:
\begin{itemize}
    \item We propose \method, which produces animations with arbitrary subject types, arbitrary counts, arbitrary spatial positions, and arbitrary poses, given a single pose sequence that is not aligned with the reference image. To the best of our knowledge, \method is the first approach to simultaneously achieve all four ``arbitrary'' properties in character image animation.
    \item Our Unbind-Rebind strategy systematically decouples pose from overly rigid spatial bindings in the reference image and rebinds control through semantic and spatial cues, thereby achieving robustness under misalignment.
    \item On our proposed \benchmark and current Follow-Your-Pose-V2 benchmark~\cite{multidance1}, \method achieves SOTA performance across metrics and shows strong generalization to diverse subjects and spatial layouts.
\end{itemize}

\section{Related Work}
\label{sec:related_work}
\subsection{Diffusion for Video Generation}
Diffusion models~\cite{DDIM,DDPM} have become the dominant engine for video synthesis~\cite{tune-a-video,ceylan2023pix2video,guo2023animatediff,zhou2022magicvideo,an2023latent,xing2023simda,qing2023hierarchical,yuan2023instructvideo, yang2024leveraging,shi2025motionstone, gong2024check, longcatvideo} thanks to their breadth of generative coverage and strong visual fidelity. Compared with images~\cite{GLIDE,Dalle2,T2i-adapter,wu2025dipo,controlnet,liu2023survey,yang2024few,peng2025mos,ji2025layerflow,ji2025physmaster,ji2025memflow}, videos add a trajectory constraint, where frames must evolve smoothly and remain temporally consistent while preserving per-frame quality. Two families of approaches have emerged. One retrofits powerful text-to-image backbones with lightweight temporal adapters~\cite{modelscopet2v, make-a-video,tft2v}, such as temporal attention blocks~\cite{guo2023animatediff} or 3D convolutions~\cite{chai2023stablevideo}, to couple appearance and motion without retraining from scratch. The other~\cite{cogvideo, tan2024mimir, hunyuanvideo} abandons the U-Net in favor of transformer~\cite{dit} diffusion backbones, which scale more gracefully to long sequences and complex dynamics. Wan2.1/2.2~\cite{wan2025} is an open Diffusion-Transformer, emphasizing long-horizon coherence and efficient inference across T2V/I2V/TI2V. Inspired by its strong empirical motion fidelity, we adopt Wan 2.1 as our backbone.

\subsection{Character Image Animation}
Character image animation targets motion transfer from a driving source to a target identity~\cite{mimicmotion2024, chang2023magicpose,tan2025SynMotion,wang2024towards,ma2025phys,peng2025stylized}, and the field has evolved markedly in both fidelity and breadth. Early approaches~\cite{li2019dense, siarohin2019first, siarohin2021motion, zhao2022thin, tan2024edtalk, tan2024flowvqtalker,ye1,ye2,ye3,ji1,ji2, tan2024say, tan2023emmn,tan2025fixtalk,tan2025edtalkpp} relied on GAN-based pipelines, which can render plausible frames yet often suffer from texture distortion, identity drift, and temporal flicker under challenging motions. The diffusion era shifted the focus to stronger priors and controllable conditioning~\cite{shen2024advancing, zhu2024champ,wu2024lamp}. Disco~\cite{disco} couples diffusion with ControlNet-style~\cite{controlnet} keypoint guidance to drive human dance; MagicAnimate~\cite{magicanimate} and Animate Anyone~\cite{Animateanyone} introduce transformer-like temporal attention to enforce cross-frame coherence and suppress jitter; UniAnimate~\cite{wang2024unianimate} leverages Mamba~\cite{mamba} for efficient long-range temporal dependencies; StableAnimator~\cite{tu2025stableanimator, tu2025stableanimator++} strengthens identity retention via a dedicated face encoder and ID-aware adaptation for high-quality human animation; and Animate-X~\cite{animatex,tan2025animatepp} enhances motion representation to support animation of a wide variety of characters. Despite these advances, most methods are architected for a single subject, which is difficult to extend to multi-subject scenes.

\cite{multidance1} introduces an image-to-video pipeline with implicit decoupling, using optical-flow, depth-order, and reference-pose guiders to stabilize backgrounds and handle occlusions, achieving strong results mainly in two-subject cases. ~\cite{multidance2}, in a text-to-video setting, employs region masks, role-wise prompt decomposition, spatially aligned cross-attention, and multi-branch control to enable two-character generation without tuning. However, both struggle to generalize beyond two subjects, limited by rigid pose–image binding that does not scale gracefully. By contrast, our Unbind–Rebind strategy first relaxes spatial couplings to learn motion semantics, then re-introduces semantic and spatial bindings, thereby naturally extending to arbitrary subject counts and more complex layouts.

\begin{figure*}[t]
   \begin{center}
    \includegraphics[width=1\linewidth]{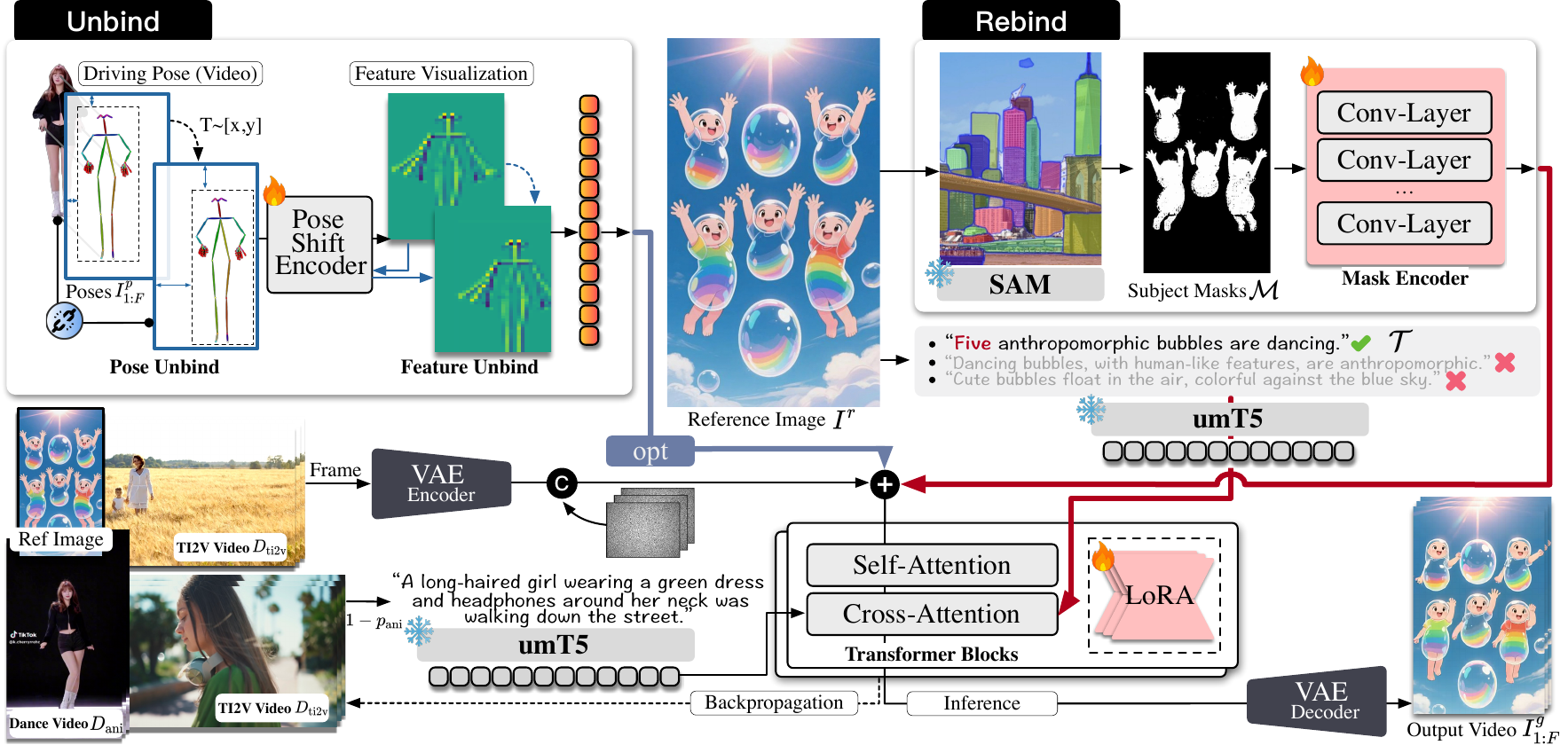}
    \end{center}
    \vspace{-3mm}
    \caption{\textbf{The pipeline of \method.} Given a reference image $I^r$, a driving pose sequence $I^p_{1:F}$, a text prompt $\mathcal{T}$, and a subject mask $\mathcal{M}$, our model generates an animation video $I^g_{1:F}$. 
    A VAE encoder extracts the latent feature $f^r_e$ from $I^r$. 
    The Unbind module, implemented as a \textbf{Pose Shift Encoder}, processes $I^p_{1:F}$ to produce pose features. 
    These are concatenated with patchified tokens from the noisy latent input for the DiT backbone. 
    The Rebind module provides dual guidance: \textbf{semantic features} from a umT5 text encoder are injected via cross-attention, while \textbf{spatial features} from a Mask Encoder are added element-wise to the noisy latent. To bolster the model's semantic comprehension, the training process alternates between animation data (with probability \(p_\text{ani}\)) and a diverse text-to-video dataset (with probability \(1-p_\text{ani}\)).
    The DiT is initialized from a pretrained T2V model and fine-tuned using LoRA. 
    Finally, a VAE decoder reconstructs the video. Note that the Unbind module and mixed-data training are applied exclusively during the training phase.}
    \label{fig:method}
\end{figure*}

\section{Methodology}
\label{sec:method}
As shown in Fig.~\ref{fig:method}, this paper focuses on multi-character animation. Given a reference image \(I^{r}\), a pose sequence $I^p_{1:F}$ and a text prompt $\mathcal{T}$, we obtain subject masks $\mathcal{M}$ for the reference image and propagate the driving motion to arbitrarily many subjects with diverse types, while maintaining identity consistency with the reference. Different from prior works focused on near-aligned inputs, we explicitly handle misalignment between $I^p_{1:F}$ and $I^{r}$, non-human / anthropomorphic characters, and multi-subject scenes.

\subsection{Preliminaries}
\label{sec:preliminary}
\paragraph{Diffusion Models.}
Diffusion Models~\cite{DDIM,DDPM} are generative models that learn to create data by reversing a noise-adding process. This involves two stages: a \textbf{forward process} that gradually adds Gaussian noise to clean data \(x_0\), and a \textbf{reverse process} that learns to remove it.

A key property is that any noisy sample \(x_t\) can be directly obtained from \(x_0\). The generative task is accomplished by training a network \(\epsilon_\theta(x_t, t, c)\) to predict the noise \(\epsilon\) from the noisy input \(x_t\), given timestep \(t\) and optional conditions \(c\). The model is optimized with objective:
\begin{equation}
    \mathcal{L}_{\text{simple}} = \mathbb{E}_{t, x_0, \epsilon} \left[ \left\| \epsilon - \epsilon_\theta(x_t, t, c) \right\|^2 \right]
    \label{eq:diffusion_loss}
\end{equation}
For conditional generation, classifier-free guidance is often used to strengthen the influence of the condition \(c\).

\paragraph{Diffusion Transformers (DiT).}
While early diffusion models used U-Net architectures, Diffusion Transformers (DiT)~\cite{dit} demonstrated that a standard Transformer can serve as a highly effective and scalable backbone. In the DiT framework, the input image is first divided into non-overlapping patches, similar to a Vision Transformer (ViT). These patches, along with embeddings for the timestep \(t\) and conditions \(c\) (e.g., pose skeletons), are converted into a sequence of tokens. This token sequence is then processed by the Transformer blocks to predict the output noise.

\subsection{Unbind-Rebind}

As illustrated in Fig.~\ref{fig:introduction}, previous methods typically enforce a rigid spatial binding between the reference image and the target pose. This paradigm can generate correct results in single-person animation, provided that the human-like reference image and target pose are spatially aligned. However, they are limited to mismatch cases, such as a different number of subjects in the reference image compared to the target pose. Due to its reliance on rigid spatial alignment, the model cannot correctly animate the subjects from the reference image. Instead, it hallucinates a new, pose-aligned person in the corresponding spatial region. To overcome this fundamental limitation, we propose a new paradigm: Unbind-Rebind, which breaks the forced spatial alignment from mismatched inputs and re-establishes the correct correspondence between motion and identity.

\paragraph{Unbind.} The Unbind module dismantles this rigid spatial constraint between the reference image and the pose. Instead of relying on simple spatial mapping, we compel our model, specifically the pose encoder and the diffusion network, to learn an abstract, semantic understanding of the motion itself. To this end, we propose a novel pose shift encoder, composed of Pose Unbind and Feature Unbind modules, which enhance the model's understanding at both the input and feature levels. The key insight is to deliberately and stochastically disrupt the natural alignment between the reference image $I^r$ and the target pose $I^p_{1:F}$ during each training step, which ensures the model cannot rely on a rigid spatial correspondence. Specifically, our Pose Unbind module operates at the input level. In each training step, we first sample a reference image $I^r$ and its corresponding driving pose $I^p$ following previous works. However, instead of feeding this pair directly to the model, we apply a series of transformations to the driving pose $I^p$. The most intuitive way to break the spatial association is by altering the pose's position and scale. Therefore, at each step, we randomly translate the skeleton's position, \textit{i.e.,} $T\sim[x,y]$, and randomly scale its size, further decoupling it from its original spatial location. 

However, Pose Unbind alone primarily strengthens the pose encoder's ability to interpret pose variations. The core generative process heavily relies on the diffusion network. To this end, we introduce the Feature Unbind module, which operates on the feature level. After the transformed pose passes through the pose encoder, we apply further augmentations to the resulting pose features. First, we apply a similar random translation. Furthermore, to force the diffusion model to adapt to various pose configurations within the feature space, we extract the feature region corresponding to the pose, randomly duplicate it, and superimpose these duplicates onto the original feature map. This process compels the diffusion model to develop a more robust semantic understanding of pose and enhances its generative capability under complex conditions.

\paragraph{Rebind.} 
Following the Unbind operation, while the model is able to grasp the semantic meaning of the motion from the pose images, it lacks the crucial information specifying the target subject for animation, as the original spatial alignment has been deliberately broken. To address this, we introduce the Rebind module, which intelligently re-associates this understood motion with the correct subjects in the reference image. Specifically, we perform Rebind through two complementary aspects: semantic and spatial. From a semantic perspective, we introduce a text-driven guidance branch that leverages an input text prompt $\mathcal{T}$ to explicitly specify the identity and number of subjects to be animated from the reference image. As shown in Fig.~\ref{fig:method},  the reference image contains multiple elements, including five anthropomorphic characters targeted for animation. A corresponding prompt, such as `\textit{Five bubbles are dancing}', is then processed by a text encoder and fed into the DiT blocks to provide semantic guidance. However, training exclusively on animation datasets (\(D_\text{ani}\)) with uniform textual prompts presents a significant challenge: the model tends to overfit to the prompt, learning a spurious correlation and ignoring the textual guidance, which severely impairs its generalization ability during inference. To counteract this, we propose a mixed-data training strategy. We incorporate an auxiliary, diverse text-image-to-video (TI2V) dataset (\(D_\text{ti2v}\)) and alternate between character animation and T2V tasks with probabilities \(p_\text{ani}\) and \(1-p_\text{ani}\), respectively. This dual-objective training compels the model to move beyond simple pattern matching and develop a robust understanding of text conditioning. This, in turn, enables it to accurately rebind specified subjects from the reference image based on arbitrary text prompts during inference.

Semantic guidance, while powerful, does not resolve the challenge of figure-ground ambiguity, particularly for subjects with complex or unconventional morphologies. This ambiguity can cause the model to fail at accurate subject segmentation, leading to outputs where background is mistakenly animated or parts of the subject are omitted. To enforce precise spatial control, we introduce a spatial rebind mechanism, which provides a reference mask $\mathcal{M}$ that explicitly defines the animation region. This direct spatial rebind ensures the animation is strictly confined to the specified boundaries, effectively mitigating segmentation errors and preserving the subject's structural integrity.

\begin{table*}[!t]
\setlength\tabcolsep{5pt}
\centering
\resizebox{0.85\linewidth}{!}{
\begin{tabular}{l|ccccc|cc}
\shline
Method        & LPIPS $\downarrow$ & PSNR $\uparrow$ & SSIM $\uparrow$ & L1 $\downarrow$  & FID $\downarrow$  &FID-VID $\downarrow$  & FVD $\downarrow$ \\ \shline

AnimateAnyone~\cite{Animateanyone}$_{\color{gray}{\text{(CVPR24)}}}$   & 0.183 & 22.08 & 0.867 & $3.15 \times E^{-5}$ & 38.30          & 26.58          & 696.43 \\
    MusePose~\cite{musepose} $_{\color{gray}{\text{(ArXiv24)}}}$        & 0.187 & 22.67 & 0.875 & $3.01 \times E^{-5}$ & 48.25          & 21.81          & 489.04 \\
    ControlNeXt ~\cite{peng2024controlnext} $_{\color{gray}{\text{(ArXiv24)}}}$    & 0.257 & 19.43 & 0.806 & $4.52 \times E^{-5}$ & 73.92          & 41.74          & 748.70 \\
    MimicMotion ~\cite{mimicmotion2024} $_{\color{gray}{\text{(ICML25)}}}$    & 0.292 & 17.99 & 0.782 & $5.20 \times E^{-5}$ & 91.05          & 68.86          & 1304.00 \\
    UniAnimate~\cite{wang2024unianimate} $_{\color{gray}{\text{(SCIS25)}}}$ & 0.183 & 21.32 & 0.837 & $2.88 \times E^{-5}$ & \textbf{35.93} & 19.60          & 366.90 \\
    Animate-X ~\cite{animatex} $_{\color{gray}{\text{(ICLR25)}}}$ & 0.162 & 24.00 & 0.884 & $2.65 \times E^{-5}$ & 41.67          & 14.81          & 392.43 \\
    StableAnimator ~\cite{tu2025stableanimator} $_{\color{gray}{\text{(CVPR25)}}}$ & 0.214 & 20.92 & 0.869 & $3.98 \times E^{-5}$ & 123.94         & 82.41          & 947.10 \\
    UniAnimateDiT~\cite{unianimatedit} $_{\color{gray}{\text{(ArXiv25)}}}$  & 0.159 & 23.59 & \textbf{0.896} & $2.59 \times E^{-5}$ & 43.86 & 20.77          & 366.19 \\


\textbf{\method}  & \textbf{0.153} & \textbf{25.76} & \textbf{0.896} & $\mathbf{2.52 \times E^{-5}}$ & 38.98 & \textbf{11.56} & \textbf{312.13}      \\

\shline
\end{tabular} 
}
\vspace{-1mm}
\caption{
Quantitative comparisons with existing methods on Follow-Your-Pose-V2~\cite{multidance1}.
}
\label{tab:quantitative_Follow-Your-Pose-V2}
\end{table*}

\begin{table*}[!t]

\renewcommand{\arraystretch}{1}
\setlength\tabcolsep{5pt}
\centering
{\begin{tabular}{l|ccccc|cc}
\shline
Method        & LPIPS $\downarrow$ & PSNR $\uparrow$ & SSIM $\uparrow$ & L1 $\downarrow$  & FID $\downarrow$  &FID-VID $\downarrow$  & FVD $\downarrow$ \\ \shline

AnimateAnyone~\cite{Animateanyone}$_{\color{gray}{\text{(CVPR24)}}}$   & 0.633 & 11.01 & 0.558 & $1.44 \times E^{-4}$ & 229.61 & 187.83 & 3414.23 \\
    MusePose~\cite{musepose} $_{\color{gray}{\text{(ArXiv24)}}}$        & 0.580 & 12.01 & 0.587 & $1.29 \times E^{-4}$ & 217.38 & 191.43 & 2952.49 \\
    ControlNeXt ~\cite{peng2024controlnext} $_{\color{gray}{\text{(ArXiv24)}}}$    & 0.652 & 10.46 & 0.355 & $1.52 \times E^{-4}$ & 222.16 & 187.27 & 3625.11 \\
    MimicMotion ~\cite{mimicmotion2024} $_{\color{gray}{\text{(ICML25)}}}$    & 0.592 & 11.63 & 0.554 & $1.34 \times E^{-4}$ & 207.22 & 182.24 & 3972.27 \\
    UniAnimate~\cite{wang2024unianimate} $_{\color{gray}{\text{(SCIS25)}}}$ & 0.582 & 11.87 & 0.582 & $1.30 \times E^{-4}$ & 207.39 & 182.27 & 3113.85 \\
    Animate-X ~\cite{animatex} $_{\color{gray}{\text{(ICLR25)}}}$ &  0.580 & 12.06 & 0.579 & $1.28 \times E^{-4}$ & 226.66 & 195.65 & 2795.78 \\
    StableAnimator ~\cite{tu2025stableanimator} $_{\color{gray}{\text{(CVPR25)}}}$ &0.604 & 12.18 & 0.575 & \textbf{1.24} $\times E^{-4}$ & 219.06 & 196.99 & 3041.96 \\
    UniAnimateDiT~\cite{unianimatedit} $_{\color{gray}{\text{(ArXiv25)}}}$  & \textbf{0.579} & 12.07 & 0.588 & $1.28 \times E^{-4}$ & \textbf{205.30} & 190.79 & 2825.82 \\

\textbf{\method}  & {0.580 }& \textbf{12.21} & \textbf{0.592} & \textbf{1.24 $\times E^{-4}$ }& 221.40 & \textbf{180.50} & \textbf{2494.76} \\

\shline
\end{tabular} 
}
\vspace{-1mm}
\caption{
Quantitative comparisons with existing methods on \benchmark.
}
\vspace{-5mm}
\label{tab:quantitative_ourbench}
\end{table*}

\subsection{Framework and Implement Details}
\label{sec:codance}
In light of the success of previous works~\cite{unianimatedit, wan2025}, \method is based on the commonly used Diffusion Transformer (DiT)~\cite{dit}. As shown in Fig.~\ref{fig:method}, given a reference image $I^r$, we employ the VAE encoder to extract its latent representation $f^r_e$. Following~\cite{unianimatedit}, this latent is then directly used as part of the input for the denoising network $\epsilon_\theta$. To facilitate a precise appearance rebind, we utilize a pretrained segmentation model (e.g., SAM) to extract a corresponding subject mask $\mathcal{M}$ from $I^r$. This mask is further fed into a Mask Encoder, consisting of stacked 2D convolutional layers. The resulting mask features are then incorporated by element-wise summation with the noisy latent vector. Concurrently, we introduce a umT5 Encoder~\cite{t5} for semantic understanding. The text features are integrated into the generative process via the cross-attention layers within the DiT blocks. For the driving video $I^p_{1:F}$, we adopt the introduced Pose Shift Encoder, composed of multiple stacked 3D convolutional layers, to effectively capture both temporal and spatial features of the driving poses. The extracted pose features are concatenated with the patchified image tokens. These conditioned tokens are then fed into our DiT backbone. Leveraging the exceptional text-to-video (T2V) generation capabilities of the pretrained Wan2.1 14B model, we initialize our DiT with its weights and keep them frozen during training. Fine-tuning is performed exclusively on newly introduced Low-Rank Adaptation (LoRA) layers integrated within the self-attention and cross-attention blocks. Finally, the denoised latent representation is passed to the VAE decoder to reconstruct the final pixel-level video $I^g_{1:F}$. It is important to note that both the Unbind modules and the mixed-data training strategy are applied only during the training phase. They are entirely bypassed during inference, ensuring no additional computational overhead.

\section{Experiments}
\label{sec:exp}
\subsection{Experimental Settings}

\begin{figure*}[t]
   \begin{center}
    \includegraphics[width=1\linewidth]{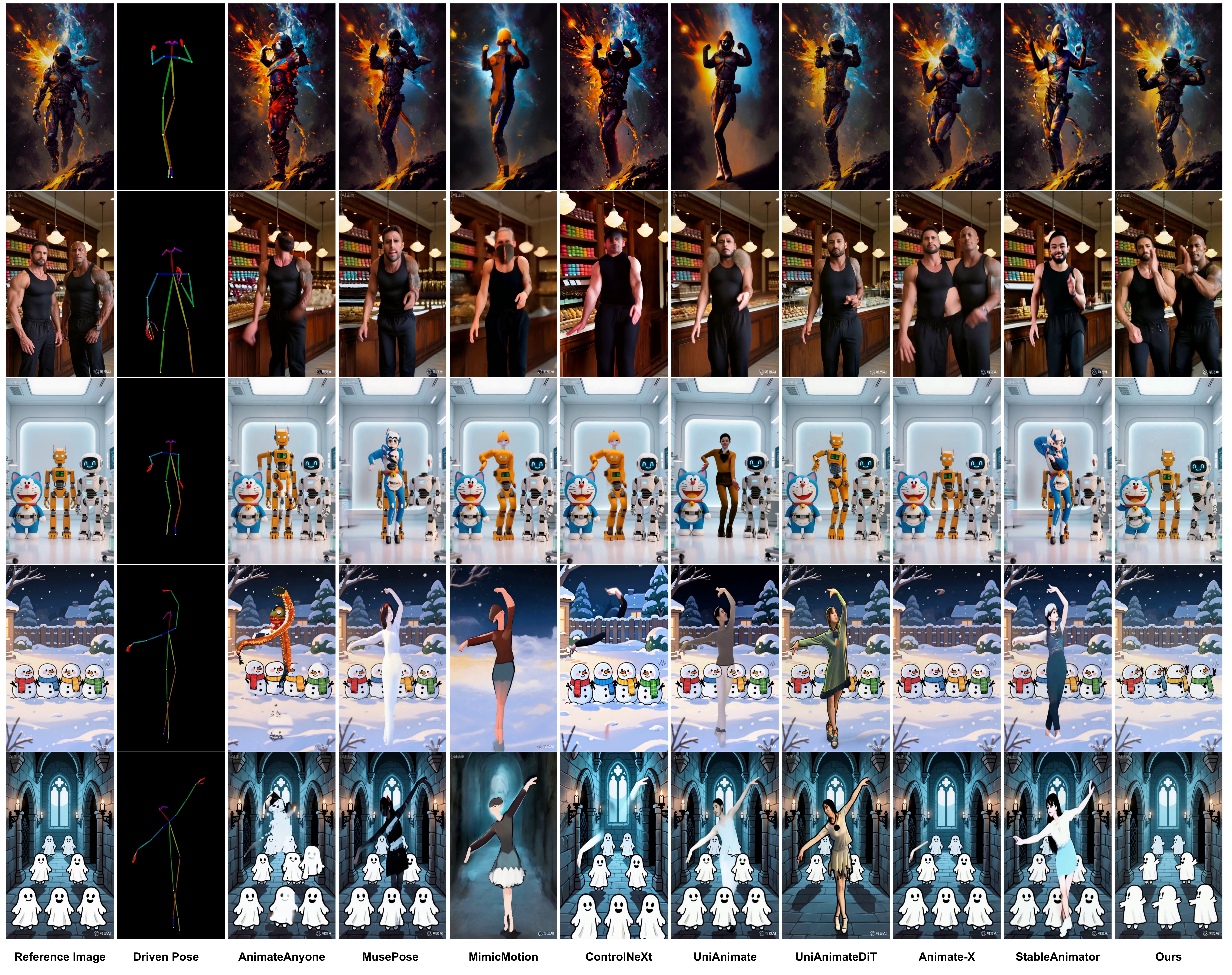}
    \end{center}
    \vspace{-7mm}
    \caption{Qualitative comparisons with SOTA methods.}
    \label{fig:compare_1}
    \vspace{-5mm}
\end{figure*}

\noindent \textbf{Dataset.} Our training data is composed of the public TikTok~\cite{tiktokdata} and Fashion~\cite{UBCfashion} datasets, supplemented by approximately 1,200 self-collected TikTok-style videos. During the Rebind stage, this training is augmented with 10,000 text-to-video samples to enhance semantic association and 20 multi-subject videos to supervise spatial binding. For evaluation, we assess our method on both single- and multi-subject settings. In the single-person context, we follow prior works~\cite{Animateanyone,UBCfashion,tiktokdata} by conducting qualitative and quantitative comparisons on 10 selected videos from TikTok and 100 from the Fashion dataset. For multi-subject animation, we utilize the established Follow-Your-Pose-V2 benchmark~\cite{multidance1} and additionally introduce our own curated benchmark, \benchmark, which consists of 20 multi-subject dance videos. Notably, for the quantitative results reported in this paper, we ensure a fair comparison with existing methods by training our model exclusively on solo dance videos, without incorporating the multi-subject data from our proposed \benchmark.

\noindent \textbf{Evaluation Metrics.}
We evaluate performance using the metrics defined in Appendix. For frame-level quality, we report PSNR \cite{hore2010image}, SSIM \cite{wang2004image}, L1, and LPIPS \cite{zhang2018unreasonable}, which are standard indicators of perceptual fidelity and distortion. To assess distributional realism at the video level, we further measure FID \cite{heusel2017gans}, FID-VID \cite{balaji2019conditional}, and FVD \cite{unterthiner2018towards}, capturing the gap between synthesized and real video distributions.

\subsection{Experimental Results}
\noindent \textbf{Quantitative Results.}
To validate the effectiveness of our proposed method, we compare it with existing SOTA approaches, including AnimateAnyone~\cite{Animateanyone}, MusePose~\cite{musepose}, ControlNeXt~\cite{peng2024controlnext}, MimicMotion~\cite{mimicmotion2024}, UniAnimate~\cite{wang2024unianimate}, Animate-X~\cite{animatex}, StableAnimator~\cite{tu2025stableanimator}, and UniAnimateDiT~\cite{unianimatedit}. Although Follow-Your-Pose-V2~\cite{multidance1} and Follow-Your-MultiPose~\cite{multidance2} address two-person animation, they are excluded from comparison as their code is not available.

The evaluated methods are fundamentally designed for single-person animation and operate on the assumption of a one-to-one correspondence between reference subjects and driving skeletons. To assess their applicability in multi-subject contexts, we conduct evaluations on the Follow-Your-Pose-V2~\cite{multidance1} dataset and our proposed benchmark. In this setup, the full set of pose skeletons from the driving video is provided as input, forcing these models into a multi-agent scenario they are not designed for. As demonstrated in Tab.~\ref{tab:quantitative_Follow-Your-Pose-V2} and Tab.~\ref{tab:quantitative_ourbench}, our method significantly outperforms all competing methods across key metrics for perceptual similarity (LPIPS), identity consistency (PSNR/SSIM), and motion fidelity (FID-FVD and FVD). These results underscore the inherent inability of conventional single-person architectures to manage multi-agent dynamics, leading to predictable failures such as identity confusion, visual artifacts, and motion distortion. Conversely, our Unbind-Rebind strategy enables our model to preserve distinct identities and generate coherent motion for each character. Critically, this robustness is achieved despite our model being trained under the same single-person data constraint as the baselines.

However, a frequent yet challenging scenario arises when a multi-subject reference image must be animated by a single-person driving pose. This creates a significant cardinality mismatch between the visual reference and the motion input. To evaluate model performance under this condition, we devise a new protocol using our benchmark. For each case, a single skeleton is isolated from the original multi-subject driving video to serve as the sole motion driver. The models are then tasked with animating the multi-subject reference image using only this single-person pose. Crucially, the generated output is quantitatively evaluated against the ground-truth multi-subject video. The detailed results for this challenging setting are provided in the supplementary material.

\begin{table*}[t]
\renewcommand{\arraystretch}{1}
\setlength\tabcolsep{2pt}
\centering
\resizebox{\linewidth}{!}{
\begin{tabular}{l|ccccccccc} 
\shline
Method         &   {AnimateAnyone} & {MusePose} & {MimicMotion} & {ControlNeXt} & {UniAnimate} & {Animate-X} & {StableAnimator} & {UniAnimateDiT} & \method  \\ \shline

Video Quality $\uparrow$ & 0.06 & 0.37 & 0.50 & 0.29 & 0.60 & 0.62 & 0.38 & 0.79 & \textbf{0.90} \\
Identity Preservation $\uparrow$    & 0.33 & 0.30 & 0.08 & 0.65 & 0.54 & 0.96 & 0.27 & 0.50 & \textbf{0.88} \\
Temporal Consistency   $\uparrow$    & 0.01 & 0.39 & 0.68 & 0.25 & 0.65 & 0.48 & 0.44 & 0.78 & \textbf{0.83} \\

\shline
\end{tabular} 
}
\vspace{-3mm}
\caption{
User study results.
}
  \vspace{-2mm}
\label{tab:user_study}
\end{table*}

\noindent \textbf{Qualitative Results.}
Qualitative comparisons are showcased in Fig.~\ref{fig:compare_1}. To evaluate generalization across varying numbers of subjects, our evaluation spans test cases featuring one to five subjects. In the standard single-person setting, even minor spatial misalignments between the reference image and the driving pose (\textit{e.g.}, differing aspect ratios or body shapes) cause competing methods to exhibit severe artifacts, such as shape distortion and identity loss. This failure is exacerbated in multi-subject scenarios. 
Constrained by their rigid binding mechanism (as illustrated in Fig.~\ref{fig:introduction}), these methods tend to overlook essential reference information, particularly the identity and number of subjects present in the scene. When driven by a single skeleton pose map, they fail to correctly associate motion cues with the intended subjects, which often leads to visually inconsistent or semantically incorrect outputs.
Animate-X presents a notable exception, where its enhanced motion representation, derived from a pose indicator, allows it to better preserve subject shape in single-person cases. However, it lacks a mechanism to rebind this global motion to specific individuals. Consequently, in multi-subject scenes, it treats all subjects as a monolithic entity, animating them in unison rather than assigning distinct motions. By contrast, our method 
successfully preserves the identity of each subject while assigning the correct, corresponding motion. The consistent generation of accurate and coherent results, irrespective of subject count, demonstrates the effectiveness and robustness of our proposed Unbind-Rebind approach.

\begin{figure}[t]
   \begin{center}
    \includegraphics[width=1\linewidth]{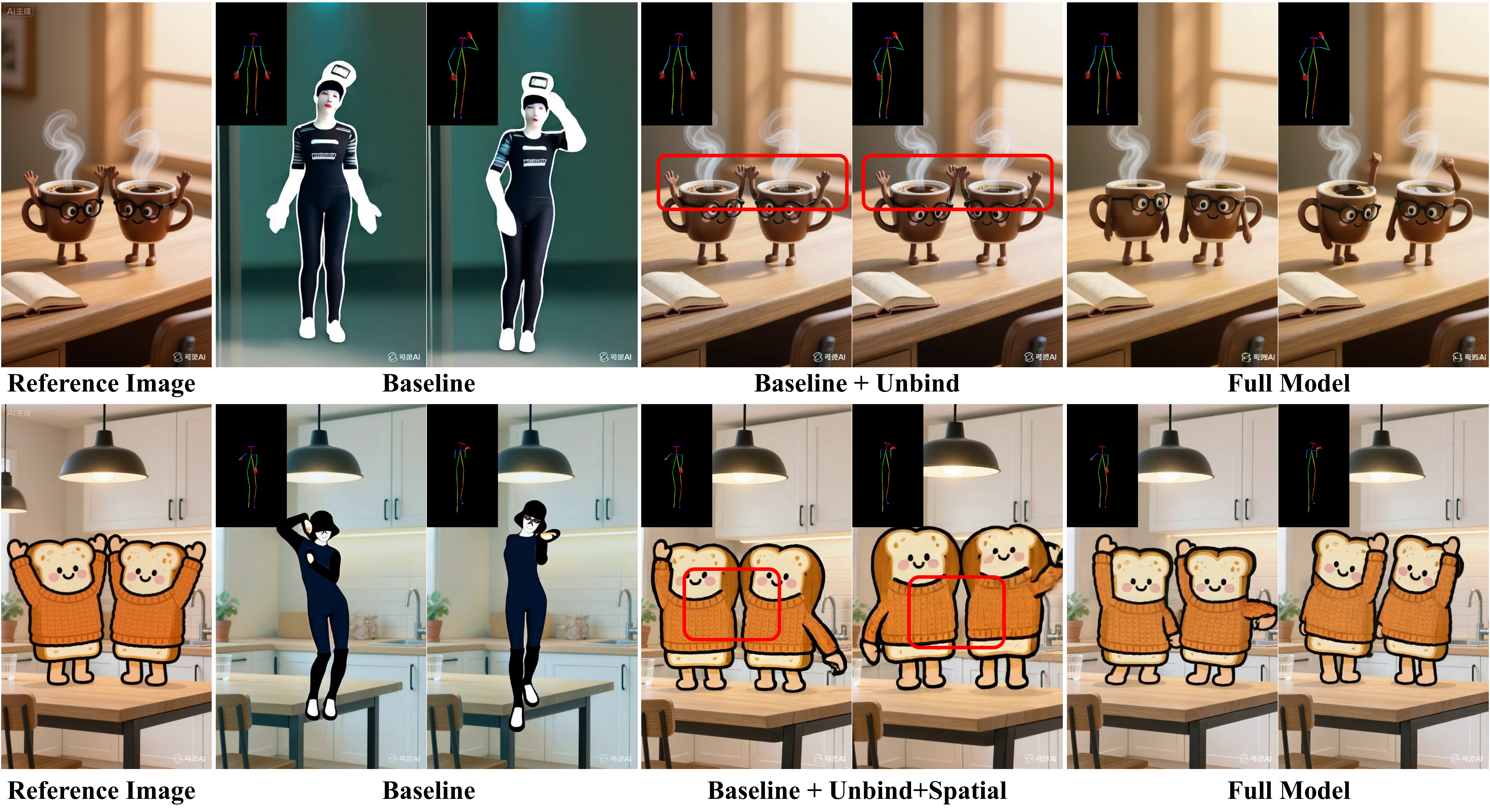}
    \end{center}
    \vspace{-5mm}
    \caption{Qualitative results of ablation study.}
    \label{fig:ablation_study}
    \vspace{-5mm}
\end{figure}

\noindent \textbf{User Study.} 
To quantify the perceptual quality, we conduct a comprehensive user study, which involves a pairwise A/B preference test administered to 10 participants. A diverse set of 20 identities and 20 driving videos is used to generate 20 animations from each of the 9 evaluated methods. In each trial, participants are presented with two side-by-side videos generated by different methods and then asked to select the superior result based on three criteria: (1) video quality, (2) identity preservation, and (3) temporal consistency. 
As summarized in Tab.~\ref{tab:user_study}, our method achieves the highest preference rates across all three criteria, demonstrating its clear perceptual superiority. 

\subsection{Ablation Study}
This section presents an ablation study designed to isolate the contribution and necessity of the Unbind and Rebind modules in \method. We structure our experiments as a progressive ablation: (1) Baseline: We remove both the Unbind and Rebind modules. The model is trained to animate the reference image following a rigid alignment paradigm following~\cite{unianimatedit}. (2) B + Unbind: On top of the Baseline, we add the Unbind module to break the rigid alignment between the reference image and the driving pose.
(3) B + Unbind + Spatial Rebind: Building on (2), we incorporate the mask condition to perform spatial rebinding. 
(4) Full Model. As illustrated in Fig.~\ref{fig:ablation_study}, the Baseline, constrained by rigid alignment, synthesizes a novel character spatially aligned with the driving pose, thereby discarding the reference identity. The introduction of the Unbind module rectifies this, preserving the reference identity and demonstrating a successful decoupling from rigid alignment. However, it fails to generate coherent motion, indicating an inability to localize the target region for animation. Adding Spatial Rebind resolves this localization issue, animating the correct regions. However, it treats multiple subjects as a single composite entity, leading to fragmented animations (e.g., animating a hand from each character instead of one character's full body). Finally, the full model, which integrates both Unbind and the complete Rebind mechanism, achieves superior results. This progression validates the crucial and complementary roles of each proposed module in our framework.
\section{Conclusion}

In this paper, we introduce \method, a novel framework engineered for robust animation across arbitrary subject counts, types, and spatial layouts. We identify that the identity degradation and motion misassignment prevalent in multi-subject scenarios stem from the rigid spatial binding in existing methods. To overcome this, we propose the Unbind-Rebind paradigm, which first unbinds motion from its strict spatial context, and then rebinds this motion to the correct subjects using complementary semantic and spatial guidance. In this way, \method demonstrates strong generalization and robustness, achieving flexible multi-subject animation. Our proposed method showcases improvements over SOTA methods, as evidenced by extensive experiments on both the Follow-Your-Pose-V2 benchmark and our newly introduced \benchmark. For limitations and future work, please see the Appendix.

{
    \small
    \bibliographystyle{ieeenat_fullname}
    \bibliography{main}
}

\end{document}